\documentclass{llncs}
\usepackage{a4,a4wide}

\usepackage{hyperref}
\usepackage{times}
\usepackage{url}
\usepackage{subfigure}
\usepackage{tikz,pgf}

\usepackage{amssymb,amsfonts,amsmath}

\tikzstyle{arg}=[draw, thick, circle]
\usetikzlibrary{arrows}

\newcommand{\NP}{\mbox{\rm NP}}

\newcommand{\PiP}[1]{{\rm \Pi}_{#1}^{P}}

\newcommand{\T}{{\ensuremath{\cal T}}}

\newcommand{\toolname}{\emph{dynPARTIX} } %

\begin{document}
\frontmatter          %

\mainmatter              %
\title{\toolname - A Dynamic Programming Reasoner for Abstract Argumentation\thanks{%
Supported by the
Vienna Science and Technology Fund (WWTF) under
grant ICT08-028,
by the Austrian Science Fund (FWF) under grant P20704-N18,
and by the Vienna University of Technology program ``Innovative Ideas''.}%
}
\author{Wolfgang Dvo\v{r}\'ak \and Michael Morak \and Clemens Nopp \and Stefan Woltran}
\authorrunning{Dvo\v{r}\'ak et al.} %
\institute{Institute of Information Systems\\ Vienna University of Technology, Austria}

\maketitle              %

\begin{abstract}
The aim of this paper is to announce the release of a novel system
for abstract argumentation which is based on decomposition and dynamic programming.
We provide first experimental evaluations to show the feasibility of this approach.
\end{abstract}

\section{Introduction}
Argumentation has evolved as an important field in AI,
with
abstract argumentation frameworks (AFs, for short)
as introduced by %
Dung~\cite{Dung95}
being its most popular formalization. 
Several semantics for AFs have been proposed (see e.g.\ \cite{Baroni09} for an overview), 
but here we shall focus on the so-called preferred semantics.
Reasoning under this semantics is known to be intractable \cite{DunneB02}.
An interesting approach to dealing with intractable problems
comes from parameterized complexity theory which suggests to focus on parameters
that allow for fast evaluations as long as these parameters are kept small. 
One important parameter for graphs 
(and thus for argumentation frameworks) is tree-width, which measures
the ``tree-likeness'' of a graph. To be more specific, 
tree-width is defined via a certain decomposition of graphs, the so-called tree decomposition.
Recent work 
\cite{DvorakPW2010} 
describes novel
algorithms for reasoning in the preferred semantics, such that 
the performance mainly depends on the tree-width of the given AF,
but the running times remain linear in the size of the AF. 
To put this approach to practice, %
we shall use the \emph{SHARP} framework\footnote{\url{http://www.dbai.tuwien.ac.at/research/project/sharp}}, a C++ environment which includes heuristic methods 
to obtain tree decompositions \cite{Dermaku08},  provides an 
interface to run algorithms on these decompositions, and offers further useful features, for instance
for parsing the input.
For a description of the \emph{SHARP} framework, see \cite{MorakDoku}.

The main purpose of our work here is to support the theoretical results 
from \cite{DvorakPW2010} with experimental ones. Therefore we use different
classes of AFs and analyze the performance of %
our approach compared to an implementation based on answer-set programming (see~\cite{EglyGW2010}).
Our prototype system together with the used benchmark instances is available as 
a ready-to-use tool from 
\url{http://www.dbai.tuwien.ac.at/research/project/argumentation/dynpartix/}.

\section{Background}
\paragraph{Argumentation Frameworks.}
An {\em argumentation framework (AF)} is a pair $F=(A,R)$ where $A$ 
is a set of arguments and $R \subseteq A \times A$ is the attack relation. 
If $(a,b)\in R$ we say $a$ attacks $b$. 
An $a \in A$ is {\em defended} by a set 
$S \subseteq A$
iff for each $(b,a) \in R$, 
there exists a $c\in S$ such that $(c,b) \in R$.
An AF can naturally be represented as a digraph.

\begin{example}\label{example:argumentation_framework}
 Consider the AF $F=(A,R)$, with
    $A=\{a,b,c,d,e,f,g\}$ and
    $R=\{(a,b)$, $(c,b)$, $(c,d)$, $(d,c)$, $(d,e)$, $(e,g)$, $(f,e)$,$(g,f)\}$.
The graph representation of $F$ 
 is given as follows:
\begin{center}
\vspace{-8pt}
\begin{tikzpicture}[scale=1.4,>=stealth']
		\path 	node[arg](a){$a$}
			++(1,0) node[arg](b){$b$}
			++(1,0) node[arg](c){$c$}
			++(1,0) node[arg](d){$d$}
			++(1,0) node[arg](e){$e$}
			++(1,0) node[arg](f){$f$}
			++(1,0) node[arg](g){$g$};
		\path [left,->, thick]
			(a) edge (b)
			(c) edge (b)
			(d) edge (e)
			(f) edge (e)
			(g) edge (f)
			;
		\path [bend left, left, above,->, thick]
			(c) edge (d)
			(d) edge (c)
			(e) edge (g)
			;
\end{tikzpicture}
\end{center}
\vspace{-5pt}
\end{example}
We require the following 
semantical concepts: %
Let $F=(A,R)$ be an AF.  A set $S\subseteq A$ is 
(i)
{\em conflict-free} in $F$, 
  if there are no $a, b \in S$, such that $(a,b) \in R$;
(ii)
\emph{admissible} in $F$, 
  if $S$ is conflict-free in $F$ and each $a\in S$ is defended by $S$;
(iii)
a \emph{preferred extension} of $F$, 
  if $S$ is a $\subseteq$-maximal admissible set in $F$.
For the AF %
in Example \ref{example:argumentation_framework}, we get the admissible sets $\{\},\{a\},\{c\},\{d\},\{d,g\},\{a,c\}, 
\{a,d\}$, and $\{a,d,g\}$. Consequently, the preferred extensions of this framework are $\{a,c\},\{a,d,g\}$. \\

The typical reasoning problems associated with AFs are the following:
(1) 
Credulous acceptance asks whether a given argument is contained 
in at least one preferred extension of a given AF;
(2)
skeptical acceptance asks whether a given argument is contained 
in all preferred extensions of a given AF.
Credulous acceptance is $\NP$-complete, while skeptical acceptance is even harder, namely $\PiP{2}$-complete~\cite{DunneB02}.

\paragraph{Tree Decompositions and Tree-width.}
As already outlined, tree decompositions will underlie our implemented algorithms. 
We briefly recall this concept (which is easily adapted to %
AFs).
 A \emph{tree decomposition} of an undirected graph $G=(V,E)$ is a pair 
 $(\mathcal{T},\mathcal{X})$ 
 where  $\mathcal{T}=(V_\mathcal{T},E_\mathcal{T})$ 
 is a tree and $\mathcal{X}=(X_t)_{t\in V_{\mathcal{T}}}$ is a set 
 of so-called bags, which has to satisfy the following conditions:
(a) $\bigcup_{t\in V_{\mathcal{T}}} X_t = V$, 
    i.e.\ $\mathcal{X}$ is a cover of $V$;
(b) for each $v \in V$,
 	$\mathcal{T}|_{\{t \mid v \in X_t\}}$
    is connected;
(c) for each $\{v_i,v_j\} \in E$, %
$\{v_i,v_j\} \subseteq X_t$ for some
    $t\in V_\T$.
 The width of a tree decomposition is given by 
 $\max \{ |X_t| \mid t \in V_{\mathcal{T}}\}-1$. 
 The \emph{tree-width} of $G$ %
 is the minimum width over all tree decompositions of $G$.

It can be shown that our example AF has tree-width $2$
and next we illustrate a tree decomposition of width $2$:

\vspace{-2pt}
\begin{center}
\begin{tikzpicture}[scale=1.4,level/.style={level distance=20pt, sibling distance=40pt/#1}]
\tikzstyle{trd}=[rectangle,draw, rounded corners=2pt]
\node [trd] (r){$c,d$}
  child {
	node [trd] (l1){$b,c$}
	child {
		node  [trd] (l2){$a,b$}
    	}
    }
  child {
	node [trd](r1){$d,e$}
	child {
		node [trd](r2){$e,f,g$}
    	}
    }
  ;
\end{tikzpicture}
\end{center}

Dynamic programming algorithms traverse such tree decompositions 
(for our purposes we shall use so-called normalized decompositions, however)
and compute local solutions for each node in the decomposition. Thus the combinatorial explosion is 
now limited to the size of the bags, that is, to the width of the given tree decomposition.
For the formal definition of the algorithms, we refer to \cite{DvorakPW2010}.

\section{Implementation and \emph{SHARP} Framework}
\label{sec:implementationandsharpframework}

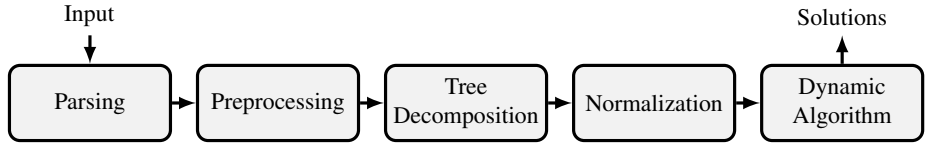
\begin{figure}[t!]
\centering
\begin{tikzpicture}[node distance = 25mm]
\tikzstyle{every path}=[very thick,draw=black,>=latex]
\tikzset{box/.style={rectangle,minimum size=10mm,text width = 20mm,
      very thick,draw=black, rounded corners, text badly centered,
      fill=black!5,inner sep=2pt}}
\node (a) [box] {Parsing};
\node (in) at (a) [above=9mm] {};
\node at (a) [above=9mm,right=4.5mm,anchor=south east] {Input};
\node (b) [right of=a,box] {Preprocessing};
\node (c) [right of=b,box]{Tree Decomposition};
\node (d) [right of=c,box]{Normalization};
\node (e) [right of=d,box]{Dynamic Algorithm};
\node (out) at (e) [above=9mm] {Solutions};
\draw [->] (in) -- (a) node {};
\draw [->] (a) -- (b);
\draw [->] (b) -- (c);
\draw [->] (c) -- (d);
\draw [->] (d) -- (e);
\draw [->] (e) -- (out);
\end{tikzpicture}
\caption{Architecture of the \emph{SHARP} framework.}
\label{fig:architectureofsharp}
\vspace{-5pt}
\end{figure}

\toolname 
implements
these algorithms %
using the
\emph{SHARP} framework %
\cite{MorakDoku}, which is a purpose-built framework
for implementing algorithms that are based on tree decompositions.
Figure \ref{fig:architectureofsharp} shows the typical 
architecture, that systems working with the \emph{SHARP} framework follow.
In fact,
\emph{SHARP} provides interfaces and helper methods
for
the Preprocessing and Dynamic Algorithm steps as well as ready-to-use
implementations of various tree decomposition heuristics,
i.e. Minimum-Fill, 
Maximum-Cardinality-Search and 
Minimum-Degree heuristics (cf.\ \cite{Dermaku08}).

\toolname builds on normalized tree decompositions provided by \emph{SHARP}, which 
contain four %
types of nodes: Leaf-,
Branch-, Introduction- and Removal-nodes.
To implement our algorithms we just have to 
provide the methods and data structures for each of these node types
(see \cite{DvorakPW2010} for the formal details). In short, the tree decomposition
is traversed in a bottom-up manner, where at each node a table of all possible partial
solutions is computed. Depending on the node type, it is then modified accordingly and
passed on to the respective parent node. 
Finally one can obtain the complete solutions from the root node's table.
\emph{SHARP} handles data-flow management and
provides data structures where the calculated (partial) solutions to the
problem under consideration can be stored. The amount of dedicated code
for \toolname\ comes to around 2700 lines in C++. Together with the  \emph{SHARP}
framework (and the used libraries for the tree-decomposition heuristics), 
our system roughly comprises of 13 000 lines of C++ code.

\section{System Specifics}\label{features}
Currently the implementation is able to calculate the admissible and preferred extensions of the given 
argumentation framework and to check if credulous or skeptical acceptance holds for a specified argument.
The basic usage of \toolname is as follows:

\vspace{-0.15cm}
\begin{verbatim}
> ./dynpartix [-f <file>] [-s <semantics>] 
        [--enum | --count | --cred <arg> | --skept <arg>]
\end{verbatim}
\vspace{-0.15cm}
The argument \verb+-f <file>+ specifies the input file, %
the argument \verb+-s <semantics>+ selects the semantics to reason with, i.e. either admissible or preferred, 
and the remaining arguments choose one of the reasoning modes. 

\paragraph{Input file conventions:} We borrow the input format from the \emph{ASPARTIX} %
system~\cite{EglyGW2010}.
\toolname thus handles text files where an argument $a$ is encoded as {\tt arg(a)} and an attack $(a,b)$ is encoded as {\tt att(a,b)}. For instance, consider the following encoding of our running example 
and let us assume that it is stored in a file {\tt inputAF}.

\vspace{-0.20cm}
\begin{verbatim}
arg(a). arg(b). arg(c). arg(d). arg(e). arg(f). arg(g).
att(a,b). att(c,b). att(c,d). att(d,c). 
att(d,e). att(e,g). att(f,e). att(g,f). 
\end{verbatim}
 
\paragraph{Enumerating extensions:}  First of all, \toolname can be used to compute extensions, 
i.e. admissible sets and preferred extensions. 
For instance to compute the admissible sets of our running example 
one can use the following command:

\vspace{-0.25cm}
\begin{verbatim}
 > ./dynpartix -f inputAF -s admissible
\end{verbatim}

\paragraph{Credulous Reasoning:} 
\toolname decides credulous acceptance using proof procedures for admissible sets (even if one reasons with preferred semantics) to avoid unnecessary computational costs.
The following statement decides if the argument $d$ is credulously accepted in our running example.

\vspace{-0.25cm}
\begin{verbatim}
 > ./dynpartix -f inputAF -s preferred --cred d
\end{verbatim}
\vspace{-0.20cm}
Indeed the answer would be \textit{YES} as $\{a,d,g\}$
is a preferred extension.

\paragraph{Skeptical Reasoning:}
To decide skeptical acceptance, \toolname uses proof procedures for preferred extensions which usually results in higher computational costs (but is unavoidable due to complexity results).
To decide if the argument $d$ is skeptically accepted, the following command is used:

\vspace{-0.25cm}
\begin{verbatim}
 > ./dynpartix -f inputAF -s preferred --skept d
\end{verbatim}
\vspace{-0.20cm}
Here the answer would be \textit{NO} as $\{a,c\}$
is a preferred extension not containing $d$.

\paragraph{Counting Extensions:}
Recently the problem of counting extensions has gained some interest~\cite{BaroniDG10}. We note that our algorithms allow counting without an explicit
enumeration of all extensions (thanks to the particular nature of 
dynamic programming; see also \cite{SamerS07}).
Counting %
preferred extensions with \toolname %
is done by

\vspace{-0.25cm}
\begin{verbatim}
 > ./dynpartix -f inputAF -s preferred --count
\end{verbatim}

\section{Benchmark Tests}\label{benchmarks}

In this section we compare \toolname with 
\emph{ASPARTIX}~\cite{EglyGW2010}, one of the 
most efficient reasoning tools for abstract argumentation
(for an overview of existing argumentation systems 
 see \cite{EglyGW2010}).  
For our benchmarks we used randomly generated AFs
 of low tree-width. 
To ensure that AFs are of a certain tree-width we considered random grid-structured AFs. 
In such a grid-structured AF each argument is arranged in an $n \times m$ grid and 
attacks are only allowed between neighbours in the grid (we used a 8-neighborhood here to allow odd-length cycles)%
. 
When generating the %
instances we varied the following parameters:
the number of arguments; %
the tree-width; %
and the probability that an possible attack is actually in the AF.

\begin{figure}[t]
\subfigure[Credulous Acceptance]{\label{figure:benchmarks_a}\includegraphics[width=0.49\textwidth]{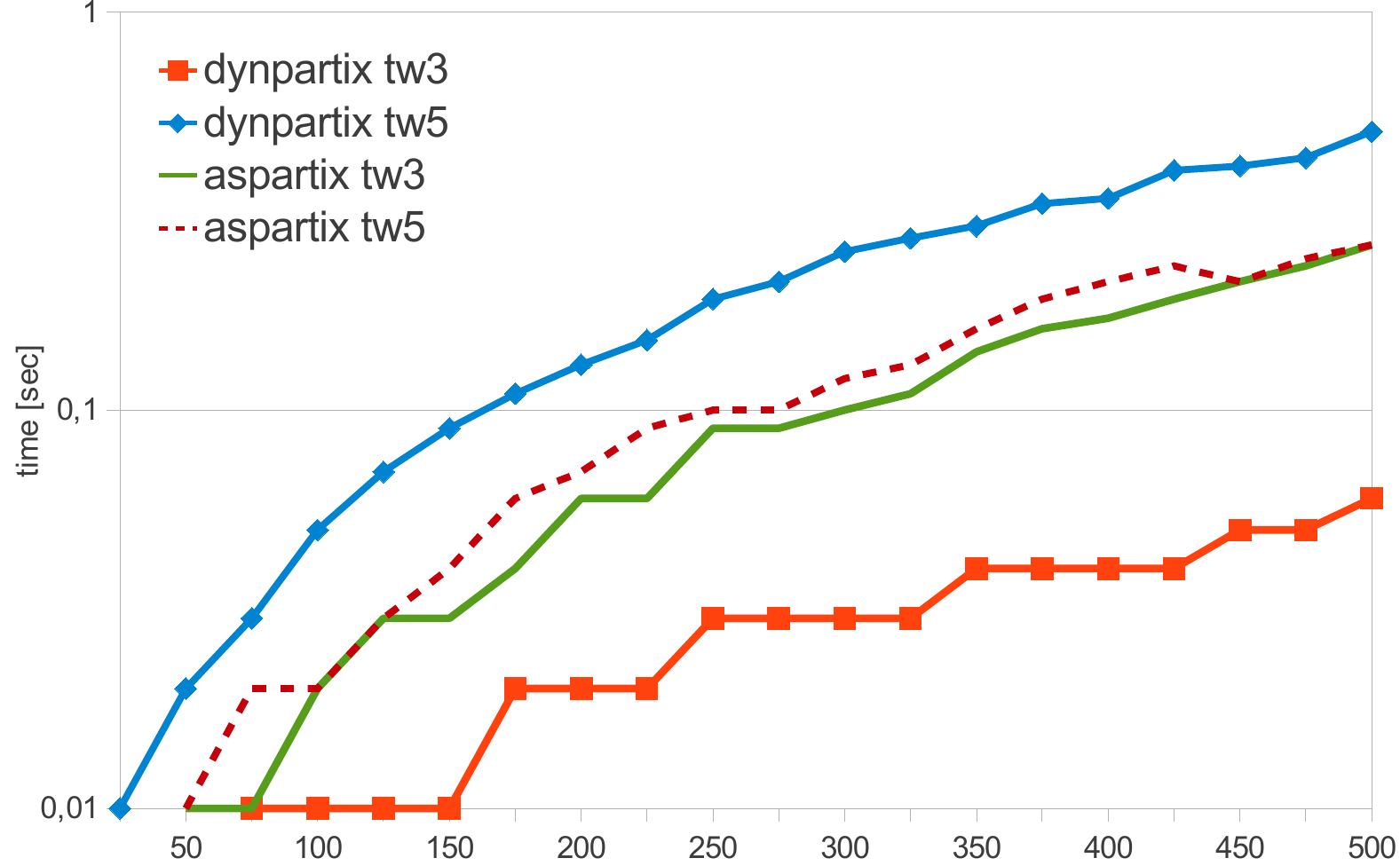}}
\subfigure[Credulous Acceptance]{\label{figure:benchmarks_b}\includegraphics[width=0.49\textwidth]{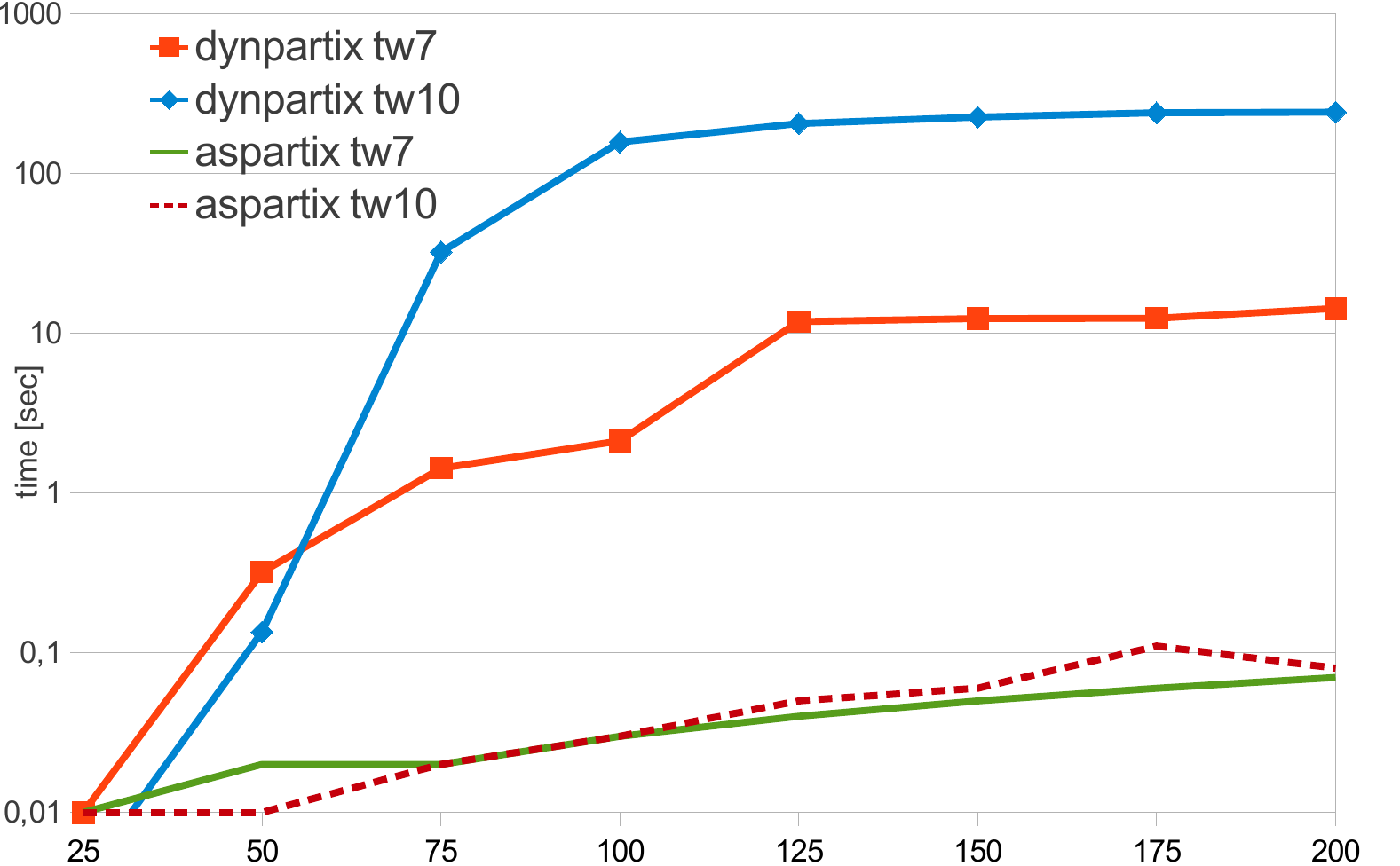}}
\subfigure[Skeptical Acceptance]{\label{figure:benchmarks_c}\includegraphics[width=0.49\textwidth]{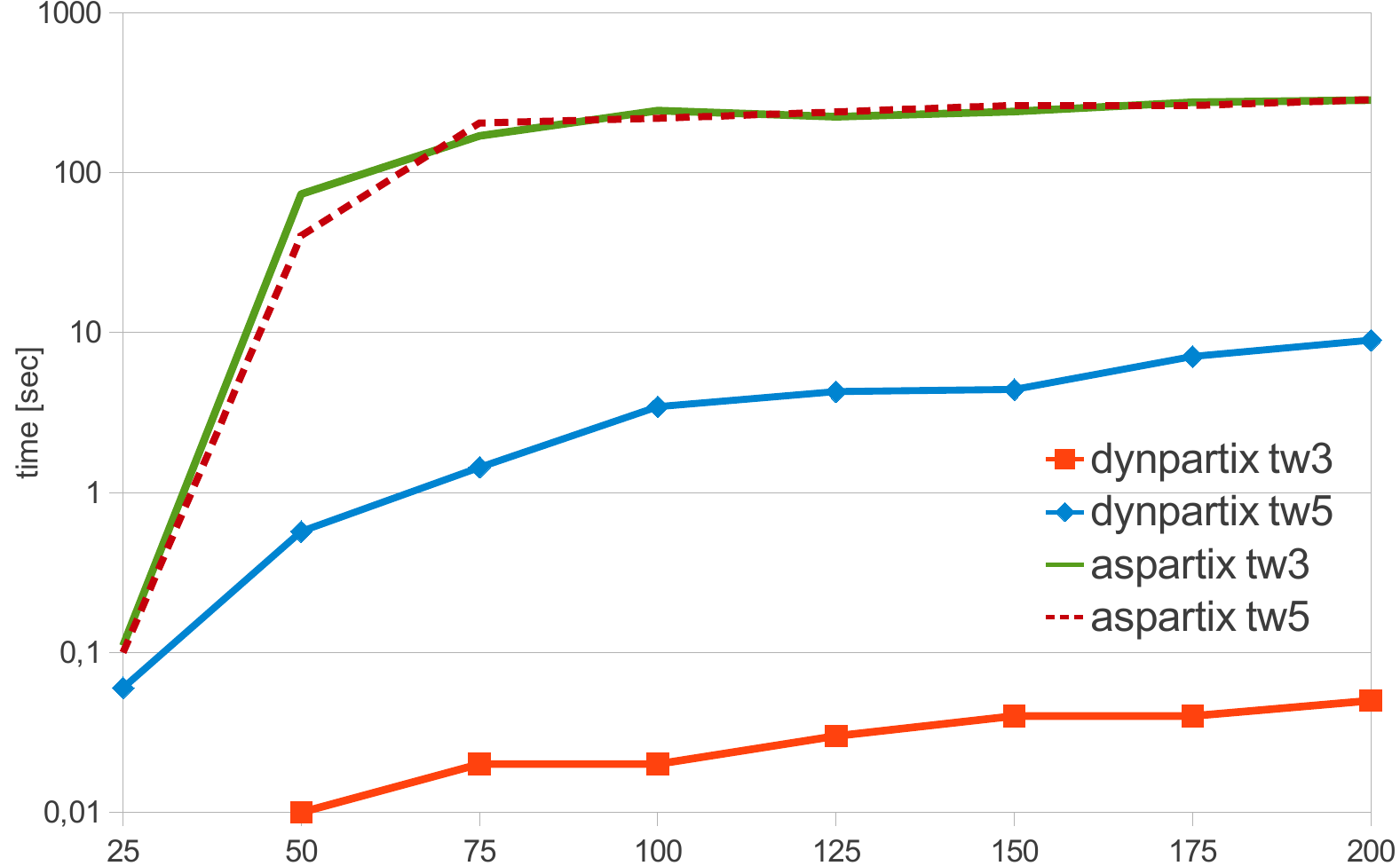}}
\subfigure[Skeptical Acceptance]{\label{figure:benchmarks_d}\includegraphics[width=0.49\textwidth]{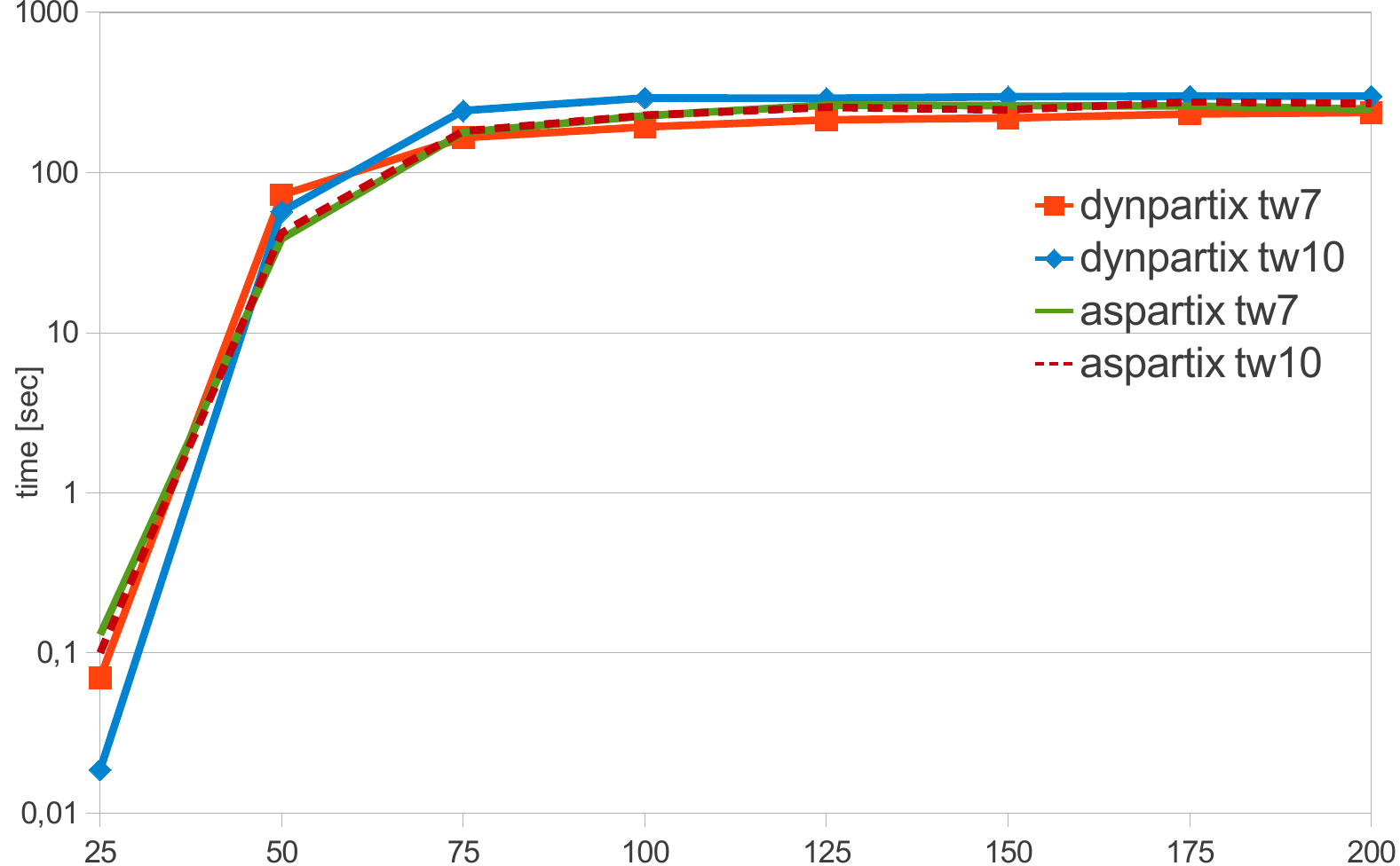}} 
\caption{Runtime behaviour of \emph{dynPARTIX} for graphs of different tree-width compared with the \emph{ASPARTIX} system.}
\label{figure:benchmarks}
\vspace{-8pt}
\end{figure}

The benchmark tests were executed on an Intel\textregistered Core\texttrademark2 CPU 6300@1.86GHz machine running SUSE Linux version 2.6.27.48. 
We generated a total of 4800 argumentation frameworks with varying parameters as mentioned above. 
The corresponding runtimes are illustrated in Figure \ref{figure:benchmarks}.
The two graphs on the left-hand side compare the running times 
of \toolname and \emph{ASPARTIX} (using dlv) on instances of small treewidth
(viz.\ 3 and 5). 
For the graphs on the right-hand side, we have used instances of higher width. 
Results for credulous acceptance are given in the upper graphs and 
those for skeptical acceptance in the lower graphs.
The y-axis gives the runtimes in logarithmic scale; the x-axis shows the number of arguments. 
Note that the upper-left picture has
different ranges on the axes compared to the three other graphs. 
We remark that the test script stopped a calculation if it was not finished after 300 seconds. For these cases we stored the value of 300 seconds in the database.

\paragraph{Interpretation of the Benchmark Results:} We observe that, independent of the reasoning mode, the runtime of \emph{ASPARTIX} is only minorly affected by the tree-width while \toolname strongly benefits from a low tree-width, as expected by theoretical results~\cite{DvorakPW2010}.

For the \emph{credulous acceptance} problem we have that our current implementation is competitive only up to tree-width $ 5$.
This is basically because \emph{ASPARTIX} is quite good at this task. Considering Figures \ref{figure:benchmarks_a} and \ref{figure:benchmarks_b}, there is to note that for credulous acceptance \emph{ASPARTIX} decided every instance in less than 300 seconds, while \toolname exceeded this value in 4\% of the cases.

Now let us consider the \emph{skeptical acceptance} problem. 
As mentioned before, skeptical acceptance is much harder computationally than credulous acceptance, 
which is reflected by the bad runtime behaviour of \emph{ASPARTIX}. %
Indeed we have that for tree-width $\leq 5$, \toolname has a significantly better runtime behaviour, and that it is competitive
on the whole set of test instances. As an additional comment to Figures \ref{figure:benchmarks_c} and \ref{figure:benchmarks_d}, we note that for skeptical acceptance, \toolname was able to decide about 71\% of the test cases within the time limit, while \emph{ASPARTIX} only finished 41\%.

Finally let us briefly mention 
the problem of \emph{Counting preferred extensions}.
On the one side we have that \emph{ASPARTIX} has no option for explicit counting extensions, 
so the best thing one can do is enumerating extensions and then counting them. %
It can easily be seen that this can be quite inefficient, which is reflected by the fact that \emph{ASPARTIX} only 
finished 21\% of the test instances in time.
On the other hand we have that the dynamic algorithms for counting preferred extensions and deciding skeptical acceptance 
are essentially the same and thus have the same runtime behaviour.

\section{Future work}

We identify several directions for future work.
First,
a  more comprehensive empirical evaluation would be of high value.
For instance, it would be interesting to explore how our algorithms perform on real world instances.
To this end, we need more knowledge about the tree-width
typical argumentation instances comprise, 
i.e.\ whether it is the case that such instances have low tree-width.
Due to the unavailability of benchmark libraries for argumentation,
so far we had to omit such considerations.

Second, we see the following directions for further development of \toolname\!:
Enriching the framework with additional argumentation semantics mentioned in \cite{Baroni09};
implementing further reasoning modes, which can be efficiently computed on tree decompositions, e.g. 
ideal reasoning; 
and
optimizing the algorithms to benefit from recent developments 
in the SHARP framework.


\begin{thebibliography}{5}

\bibitem{BaroniDG10}
Pietro Baroni, Paul~E. Dunne, and Massimiliano Giacomin.
\newblock On extension counting problems in argumentation frameworks.
\newblock In Proc. {\em COMMA 2010}, pages 63--74, 2010

\bibitem{Baroni09}
Pietro Baroni and Massimiliano Giacomin.
\newblock Semantics of abstract argument systems.
\newblock In {\em Argumentation in Artificial
  Intelligence}, pages 25--44. Springer, 2009.

\bibitem{Dermaku08}
Artan Dermaku, Tobias Ganzow, Georg Gottlob, Benjamin~J. McMahan, Nysret
  Musliu, and Marko Samer.
\newblock Heuristic methods for hypertree decomposition.
\newblock In {\em Proc.\ MICAI}, pages 1--11, 2008.

\bibitem{Dung95}
Phan~Minh Dung.
\newblock On the acceptability of arguments and its fundamental role in
  nonmonotonic reasoning, logic programming and n-person games.
\newblock {\em Artif. Intell.}, 77(2):321--358, 1995.

\bibitem{DunneB02}
Paul~E. Dunne and Trevor J.~M. Bench-Capon.
\newblock Coherence in finite argument systems.
\newblock {\em Artif.\ Intell.}, 141(1/2):187--203, 2002.

\bibitem{DvorakPW2010}
Wolfgang {Dvo\v r\'ak}, Reinhard Pichler, and Stefan Woltran.
\newblock Towards fixed-parameter tractable algorithms for argumentation.
\newblock In Proc. {\em KR'10}, pages 112--122, 2010.

\bibitem{EglyGW2010}
Uwe Egly, Sarah Gaggl, and Stefan Woltran.
\newblock Answer-set programming encodings for argumentation frameworks.
\newblock {\em In Argument and Computation}, 1(2):147--177, 2010.

\bibitem{MorakDoku}
Michael Morak.
\newblock {SHARP - A} smart hypertree-decomposition-based algorithm framework
  for parameterized problems.
\newblock Technische Universit\"{a}t Wien, 2010.\\
\newblock \url{http://www.dbai.tuwien.ac.at/research/project/sharp/sharp.pdf}

\bibitem{SamerS07}
Marko Samer and Stefan Szeider.
\newblock Algorithms for propositional model counting.
\newblock {\em J. of Discrete Algorithms}, 8(1): 50--64, 2010.

%
%
%
%
\end{thebibliography}
\end{document}